\documentclass[letterpaper,10pt,conference]{ieeeconf}  

\IEEEoverridecommandlockouts                              

\usepackage{mymacros}
\usepackage{times} 
\usepackage{graphicx}

\usepackage{caption}
\usepackage[T1]{fontenc}
\usepackage{color}
\usepackage{hyperref}
\usepackage{pifont}

\let\labelindent\relax
\usepackage{enumitem}

\usepackage{chngpage}

\usepackage{multirow}

\title{\LARGE \bf
Multimodal Diffusion Segmentation Model for \\ Object Segmentation from Manipulation Instructions
}

\author{Yui Iioka, Yu Yoshida, Yuiga Wada, Shumpei Hatanaka, Komei Sugiura}

\author{
    Yui Iioka$^{1}$, Yu Yoshida$^{1}$, Yuiga Wada$^{1}$, Shumpei Hatanaka$^{1}$, Komei Sugiura$^{1}$
    \thanks{$^{1}$The authors are with Keio University, 3-14-1 Hiyoshi, Kohoku, Yokohama, Kanagawa 223-8522, Japan.
         {\tt\small \{kmngrd1805, yu1015, yuiga, inovation, komei.sugiura\}@keio.jp}
    }%
}

\begin{document}

\maketitle
\thispagestyle{empty}
\pagestyle{empty}

\begin{abstract}

In this study, we aim to develop a model that comprehends a natural language instruction (e.g., ``Go to the living room and get the nearest pillow to the radio art on the wall'') and generates a segmentation mask for the target everyday object. 
The task is challenging because it requires (1) the understanding of the referring expressions for multiple objects in the instruction, (2) the prediction of the target phrase of the sentence among the multiple phrases, and (3) the generation of pixel-wise segmentation masks rather than bounding boxes.
Studies have been conducted on language-based segmentation methods; however, they sometimes mask irrelevant regions for complex sentences.
In this paper, we propose the Multimodal Diffusion Segmentation Model (MDSM), which generates a mask in the first stage and refines it in the second stage.
 We introduce a crossmodal parallel feature extraction mechanism and extend diffusion probabilistic models to handle crossmodal features.
To validate our model, we built a new dataset based on the well-known Matterport3D and REVERIE datasets.
This dataset consists of instructions with complex referring expressions accompanied by real indoor environmental images that feature various target objects, in addition to pixel-wise segmentation masks.
The performance of MDSM surpassed that of the baseline method by a large margin of +10.13 mean IoU.

\end{abstract}

\vspace{-1mm}
\section{Introduction
\label{intro}
}
\vspace{-1mm}

The need for assistance and support in daily life is increasing in today's aging society.
As a result, the shortage of home caregivers has become a social problem.
To solve this problem, domestic service robots (DSRs) are attracting attention because they can physically support care recipients\cite{yamamoto2019development}.
It would be convenient if natural language could be used to instruct DSRs to manipulate objects, which is an essential task for them.
However, DSRs' ability to understand natural language instructions is currently insufficient.

\begin{figure}[t]
    \centering
    \includegraphics[width=\linewidth]{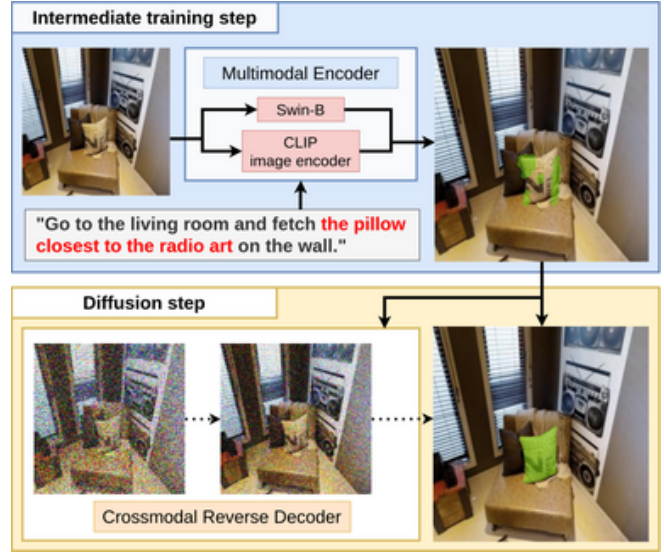}
    \caption{\small{Overview of the proposed method. Our method generates a segmentation mask for the target object of a given instruction and image in two steps. We introduce a crossmodal parallel feature extraction mechanism and extend diffusion probabilistic models to handle crossmodal features.}}
    \label{fig:eye-catch}
    \vspace{-7mm}
\end{figure}

In this study, we aim to develop a model that comprehends a natural language instruction and generate a segmentation mask for the target object.
To achieve this, we focus on the object segmentation from manipulation instructions (OSMI) task.
Fig.~\ref{fig:eye-catch} shows an example of the OSMI task.
Specifically, given the instruction ``Go to the living room and fetch the pillow closest to the radio art on the wall,''  the DSR should generate a segmentation mask for the pillow closest to the radio art.
This is because a pixel-wise segmentation mask is much more useful than a bounding box to specify the shape of the target object.

The OSMI task requires the understanding of an instruction that contains several referring expressions.
However, even comprehending a single referring expression is challenging for the DSRs.
Indeed, in \cite{qi2020reverie}, the authors reported that the accuracy of referring expression comprehension (REC) by state-of-the-art methods was approximately 50\%, whereas human performance was 90.76\%.
The OSMI task is more challenging than simple referring expression segmentation (RES) tasks because it requires (1) the understanding of the referring expressions for multiple objects in the instruction, (2) the prediction of the target object of the sentence among multiple candidates, and (3) the generation of pixel-wise segmentation masks rather than bounding boxes.

Many studies (e.g., \cite{yang2022lavt, wang2022ofa}) have been conducted on handling the RES tasks, which is closely related to the OSMI task.
In the RES tasks, the objective is to segment the object indicated by the phrase that contains a referring expression.
On the other hand, in the OSMI task, the objective is to identify one of multiple phrases that contains the referring expressions and segment the target object referred to by that phrase.
Therefore, most existing models for the RES task are insufficient for the OSMI task.

In this paper, we propose a multimodal segmentation model, the Multimodal Diffusion Segmentation Model (MDSM).
Fig.~\ref{fig:eye-catch} shows an overview of the MDSM.
The MDSM generates a segmentation mask for a target object from a natural language sentence that contains  multiple referring expressions. 
It generates a segmentation mask in the first stage and refines it in the second stage.
Unlike existing methods (e.g., \cite{baranchuklabel, ho2020denoising, dhariwal2021diffusion}), in the proposed method, we extend denoising diffusion probabilistic models (DDPM)\cite{ho2020denoising} to handle linguistic information and address the OSMI task.
Most existing methods (e.g., \cite{yang2022lavt, wang2022ofa, ding2022vlt, liu2019clevr, luo2020multi, li2021referring}) often mask irrelevant areas.
Our two-stage segmentation model masks more relevant areas than conventional single-stage models.
Furthermore, the proposed method refines relevant areas according to linguistic information.
Therefore we expect it to generate masks that appropriately reflect the instruction and object region.

We summarize the main contributions of this study as follows:
\begin{itemize}
    \item We propose the MDSM, which is a two-stage multimodal segmentation model.
    \item We introduce a crossmodal parallel feature extraction mechanism into the novel Multimodal Encoder.
    \item We extend DDPM to handle crossmodal features in the novel Crossmodal Reverse Decoder.
\end{itemize}

\begin{figure}[t]
    \centering
    \includegraphics[width=0.75\linewidth]{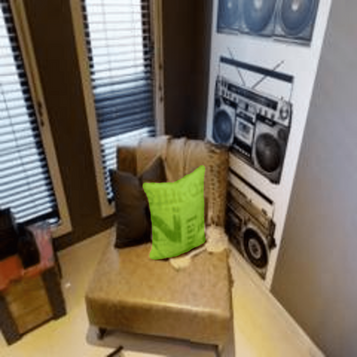}
    \caption{\small{ Typical scene for the OSMI task. The target object mask is the green highlighted region, given the instruction ``Go to the living room and fetch the pillow closest to the radio art on the wall.''}
    }
    \vspace{-3mm}
    \label{fig:task_exam}
\end{figure}

\vspace{-1.4mm}
\section{Related Work
\label{related}
}
\vspace{-0.8mm}

Many survey papers exist on vision-and-language studies\cite{UPPAL2022149, mei2020vision, chen2023vlp, qiao2020referring, ZHOU2021316, gu2022vision}.
In \cite{UPPAL2022149}, Uppal et al. presented a detailed overview of the latest trends in research on visual and language modalities.
In \cite{chen2023vlp}, the authors surveyed recent advances and new frontiers in vision-language pre-training, including image-text and video-text pre-training.

Vision-and-language studies can be divided into subfields such as REC, RES, vision-and-language navigation (VLN), and text-to-image generation.

REC aims to localize a target object in an image described by a referring expression phrased in natural language.
Wang et al. proposed an REC method in which has a unified architecture for multimodal and uni-modal understanding and generation\cite{wang2022ofa}.
Additionally, Kamath et al. proposed an end-to-end modulated detector that detects objects in an image conditioned on a raw text query, such as a caption or a question \cite{kamath2021mdetr}.

By contrast, RES refers to the pixel-wise classification task to segment the referenced area\cite{hu2016segmentation}.
LAVT\cite{yang2022lavt} is a leading method for handling RES tasks. Yang et al. showed that significantly improved crossmodal alignment can be achieved through the early fusion of linguistic and visual features in intermediate layers of a vision Transformer encoder network.
In \cite{zou2022xdecoder}, Zou et al. proposed X-Decoder, which was the first model to tackle generic and referring segmentation tasks all in one model. Furthermore, the generalized decoder jointly learns from segmentation data and image-text pairs end-to-end and thus can augment the synergy across tasks for rich pixel-level and image-level understanding.

VLN is a research area that aims to build an embodied agent that can communicate with humans in natural language and navigate in real 3D environments.
Qi et al. showed that combining instruction navigation and REC is challenging because of the large gap in human performance\cite{qi2020reverie}.

Text-to-image generation aims to generate plausible images based on text information.
Ho et al. presented high-quality image synthesis results using diffusion probabilistic models, which are a class of latent variable models based on nonequilibrium thermodynamics\cite{ho2020denoising}.
Rombach et al. transformed diffusion probabilistic models into powerful and flexible generators for general conditioning inputs such as text or bounding boxes and introduced cross-attention layers into the model architechture to make high-resolution synthesis possible in a convolutional manner\cite{Rombach_2022_CVPR}.

\begin{figure*}
    \centering
    \includegraphics[clip,width=1.\linewidth]{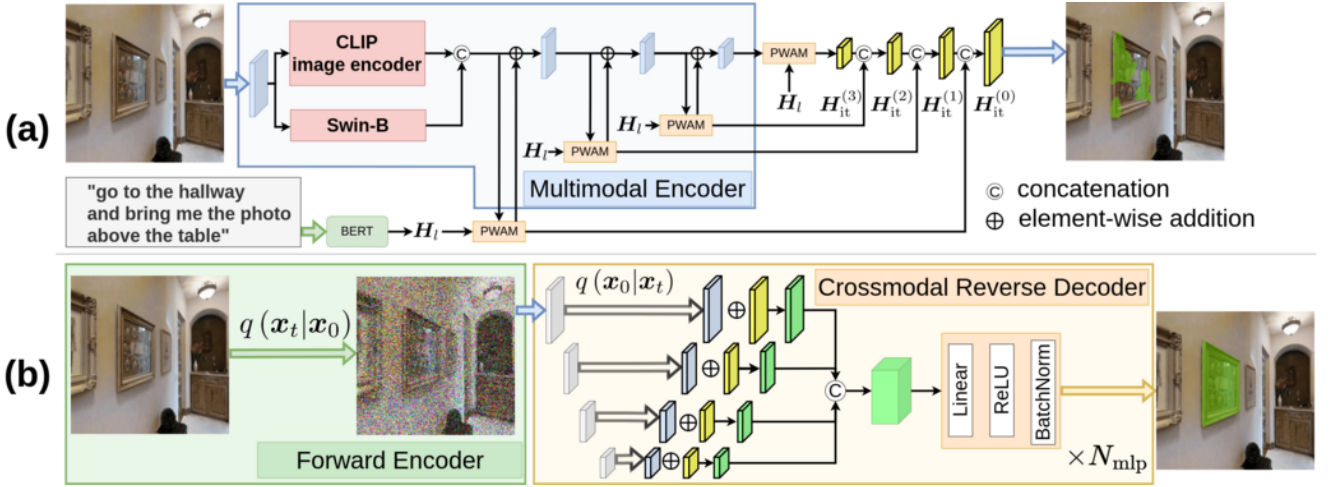}
    \caption{Framework of our method. Our method consists of two steps: (a) the intermediate training step and (b) the diffusion step. The model used in step (a) consists of two main modules, Crossmodal Encoder and the Pixel-Word Attention Module (PWAM)\cite{yang2022lavt}. The model used in step (b) consists of two main modules, Forward Encoder and Crossmodal Reverse Decoder.}
    \label{fig:model}
\end{figure*}

MTCM\cite{magassouba2019understanding} addresses the Multimodal Language Understanding for Fetching Instruction (MLU-FI) task, which requires the model to identify the target object described by the manipulation instruction, by processing the instruction in LSTM\cite{hochreiter1997long} and the image in VGG16\cite{simonyan2014very}.
In \cite{ishikawa2021target}, Ishikawa et al. proposed Target-dependent UNITER, which learned the relationship between the target object and other objects directly by focusing on relevant regions within an image rather than the entire image.
Moment-based Adversarial Training\cite{ishikawa2022moment} uses two types of  moments for perturbation updates to the embedding spaces of the instruction, subgoals, and state representations in the task that includes navigation, object interactions, and state changes.

In the fields of REC and RES, several standard datasets exits that are used to compare the methods \cite{kazemzadeh2014referitgame, yu2016modeling, mao2016generation, lin2014microsoft, interact_picking18}
The standard datasets used in RES include RefCOCO\cite{kazemzadeh2014referitgame}, RefCOCO+\cite{yu2016modeling}, and G-Ref\cite{mao2016generation}. The images in the three datasets were collected from MS COCO\cite{lin2014microsoft} and annotated with natural language expressions.
The PFN-PIC dataset\cite{interact_picking18} contains natural language instructions and corresponding object images, which were collected using a robotic system that selected and placed commodities.
Additionally, the REVERIE dataset\cite{qi2020reverie} is a dataset of household tasks in an indoor environments and contains images annotated with natural language instructions including the referring expressions for multiple objects.

Unlike existing methods (e.g., \cite{yang2022lavt, wang2022ofa, ding2022vlt, liu2019clevr, luo2020multi, li2021referring}), the proposed method, MDSM, is a two-stage multimodal segmentation model.
Furthermore, unlike LAVT\cite{yang2022lavt}, we introduce a parallel cross-modal feature extraction mechanism in MDSM.
Additionally, unlike existing methods (e.g., \cite{baranchuklabel, ho2020denoising, dhariwal2021diffusion}), we introduce a diffusion probabilistic model that handles multimodal features for MDSM to address the OSMI task.

\vspace{-1.0mm}
\section{Problem Statement
\label{sec:problem}
}
\vspace{-0.8mm}

In this study, we focus on the OSMI task in real-world household tasks.
In this task, the model should generate a segmentation mask for the target object of a given instruction.
Fig.~\ref{fig:task_exam} shows an example of the OSMI task.
If the instruction is ``Go to the living room and fetch the pillow closest to the radio art on the wall,'' the goal is to generate a segmentation mask, which is indicated by the green area.

We define the terms used in this paper as follows:
\begin{itemize}
    \item \textbf{Instruction:} a statement that requests that the robot performs various household tasks.
    \item \textbf{Target object:} the object referred to by the instruction. We assume that there is only one target object in the image.
\end{itemize}

The input and output are defined as follows:
\begin{itemize}
    \item \textbf{Input:} An image and instruction sentence.
    \item \textbf{Output:} A pixel-wise segmentation mask.
\end{itemize}

We use mean intersection over union (mIoU), overall intersection over union (oIoU), and precision, which are standard in RES tasks, as the evaluation metrics.

\vspace{-1mm}
\section{Proposed Method: MDSM
\label{method}
}

\vspace{-0.8mm}
\subsection{Novelty}
\vspace{-1mm}

Fig.~\ref{fig:model} shows the framework of our method.
Our method is inspired by existing methods for RES tasks, such as LAVT\cite{yang2022lavt}.
The novelty of our approach is as follows:
\begin{itemize}
    \item We propose the MDSM, which is a two-stage multimodal segmentation model.
    \item We introduce a crossmodal parallel feature extraction mechanism using CLIP\cite{radford2021learning} and Swin Transformer\cite{liu2021swin}.
    \item We extend DDPM\cite{ho2020denoising} to handle crossmodal features in the novel  Crossmodal Reverse Decoder.
\end{itemize}

In this study, we evaluate the performance of our method for instruction comprehension in real-life household tasks.
However, the crossmodal parallel feature extraction mechanism and vision-and-language segmentation model can be applied to RES tasks in various environments.

\subsection{Intermediate Training Step}
In the intermediate training step, the input consists of image $\bm{x}_0 \in \mathbb{R}^{H \times W \times 3}$ and instruction $\bm{x}_L$, which is a set of one-hot vectors of dimension $v \times l$. 
where $H$, $W$, $v$, and $l$ denote the image height and width, instruction vocabulary size, and max token length, respectively.

We embedded $\bm{x}_L$ into the language features $H_l \in \mathbb{R}^{C \times l}$ using a pre-trained BERT base model\cite{devlin-etal-2019-bert}, where $C$ is the number of feature dimensions in each token.

\subsubsection{Multimodal Encoder}

The first block in the Multimodal Encoder takes $\bm{x}_0$ as input and outputs image features $\bm{V}_1 \in \mathbb{R}^{H_1 \times W_1 \times C_1}$, where $H_i$, $W_i$, and $C_i$ are the height of the image, width of the image, and number of channels in the $i$-th block, respectively.

First, $\bm{x}_0$ is input in parallel into the Swin-B\cite{liu2021swin} and CLIP image encoder\cite{radford2021learning} to obtain image features $\bm{V}_{\mathrm{sw}}^{(1)} \in \mathbb{R}^{H_1 \times W_1 \times C_1}$ and $\bm{V}_{\mathrm{clp}}^{\prime (1)} \in \mathbb{R}^{C_{\mathrm{clp}}}$.
where $C_{\mathrm{clp}}$ denotes the dimension of the image features output by the CLIP image encoder.
Next, $\bm{V}_{\mathrm{clp}}^{\prime (1)}$ is reshaped to obtain $\bm{V}_{\mathrm{clp}}^{(1)} \in \mathbb{R}^{H_{\mathrm{clp}}^{(1)} \times W_{\mathrm{clp}}^{(1)} \times C_1}$. where $H_{\mathrm{clp}}^{(1)}=W_{\mathrm{clp}}^{(1)}=\sqrt{C_{\mathrm{clp}} / C_1}$. 
Then, $\bm{V}_{\mathrm{sw}}^{(1)}$ and $\bm{V}_{\mathrm{clp}}^{(1)}$ are concatenated in the channel direction.
Finally, the model performs a $3 \times 3$ convolution on this feature and outputs $\bm{V}_1$.

Then, the $i$-th Multimodal Encoder takes the multimodal feature $\bm{E}_{i-1} \in \mathbb{R}^{H_{i-1} \times W_{i-1} \times C_{i-1}}$ as input and outputs $\bm{V}_i$, where $\bm{E}_{i-1}$ is obtained from the features $\bm{F}_{i-1} \in \mathbb{R}^{H_{i-1} \times W_{i-1} \times C_{i-1}}$ extracted by the PWAM described below and $\bm{V}_{i-1}$ as follows:
\begin{align*}
  \bm{S}_{i-1} &= \mathrm{tanh}(\mathrm{Conv}(\bm{F}_{i-1})), \\
  \bm{E}_{i-1} &= \bm{F}_{i-1} \odot \bm{S}_{i-1} + \bm{v}_{i-1},
\end{align*}
where $\bm{S}_{i-1}$, $\mathrm{Conv}$ and  $\odot$ denote the $\bm{F}_{i-1}$ weight map, $1 \times 1$ convolution and element-wise multiplication, respectively.

\subsubsection{PWAM}
The Pixel-Wise Attention Module (PWAM)\cite{yang2022lavt} takes as inputs $\bm{V}_i$ and $\bm{H}_l$ and outputs multimodal features $\bm{F}_i \in \mathbb{R}^{H_i \times W_i \times C_i}$.
Unlike ViT\cite{dosovitskiy2020image}, the PWAM requires small memory usage because it does not need to compute attention weights between two image-sized spatial feature maps.
Specifically, we obtain $\bm{F}_i$ as follows:
\begin{align*}
    \bm{G}_i^{\prime} &= \mathrm{softmax}\left(C_i^{-\frac{1}{2}}\mathrm{flatten}(\omega_{iq}(\bm{V}_i)) \left(\omega_{ik}(\bm{H}_l)\right)^\top\right) \omega_{iv}(\bm{H}_l), \\
    \bm{G}_i &= \omega_{iw}\left(\mathrm{unflatten}\left(\bm{G}_{i}^{\prime \top} \right) \right), \\
    \bm{F}_i &= \omega_{if}(\omega_{im}(\bm{V}_i)) \odot \bm{G}_i),
\end{align*}
where ($\omega_{iq}$, $\omega_{ik}$, $\omega_{iv}$, $\omega_{iw}$, $\omega_{if}$, $\omega_{im}$) denotes $1 \times 1$ convolution. Additionally, ($\bm{G}_i^{\prime}$, $\bm{G}_i$) denotes the multimodal features.
In this case, ``flatten'' refers to the operation of rolling the two spatial dimensions into one dimension and ``unflatten'' refers to the opposite operation.

\subsubsection{Output}
In the intermediate training step, we obtain the final segmentation using each $\bm{F}_i$. 
Specifically, we describe the computation as follows:
\begin{align*}
    \bm{H}_{\mathrm{it}}^{(i)} &= 
                \begin{cases}
                    \bm{F}_i & i=M, \\
                    \mathrm{Conv}\left(\left[f_{\mathrm{up}}\left(\bm{H}_{\mathrm{it}}^{(i+1)}\right); \bm{F}_{i}\right]\right) & i \neq M, \\
                \end{cases} \\
    p\left(\hat{\bm{y}}_{\mathrm{it}}\right) &= \mathrm{softmax}\left(\mathrm{Conv}\left(\bm{H}_{\mathrm{it}}^{(1)}\right)\right),
\end{align*}
where $M$ denotes the number of Multimodal Encoders , $[;]$ denotes concatenation along the channel dimension, and $f_{\mathrm{up}}$ denotes upsampling via bilinear interpolation. $\hat{\bm{y}}$ is the predicted segmentation mask. 
$p\left(\hat{\bm{y}}_{\mathrm{it}}\right)$ denotes the predicted probability in the intermediate training step.
Finally, we binarize $p\left(\hat{\bm{y}}_{\mathrm{it}}\right)$ at a threshold of 0.5 and output the $H \times W$ binary prediction mask image $\hat{\bm{y}}_{\mathrm{it}}$.

\subsection{Diffusion Step}

\subsubsection{Forward Encoder}
In the diffusion step , the input consists of $\bm{x}_0$, $\bm{H}_{\mathrm{it}}^{(i)}$, and $p\left(\hat{\bm{y}}_{\mathrm{it}}\right)$.
The Forward Encoder is a mechanism for gradually adding Gaussian noise based on Markov processes, similar to the general diffusion probability model\cite{ho2020denoising}.
It takes as input $\bm{x}_0$ and output an image $\bm{x}_t \in \mathbb{R}^{H \times W \times 3}$ to which Gaussian noise is added $t$ times.
The added noise follows a probability distribution $q(\bm{x}_t | \bm{x}_{t-1})$ represented by the following normal distribution.
\begin{align*}
    q(\bm{x}_t | \bm{x}_{t-1}) := \mathcal{N}(\bm{x}_t; \sqrt{1-\beta_t} \bm{x}_{t-1}, \beta_t \bm{I}),
\end{align*}
where $\beta_t$ is the noise weight to be added the $t$-th time, and $\bm{I}$ denotes the identity matrix.

Thus, we compute $\bm{x}_t$ using $\bm{x}_{t-1}$ as follows:
\begin{align*}
    \bm{x}_t=\sqrt{1-\beta_t}\bm{x}_{t-1} + \sqrt{\beta_t} \bm{\epsilon}, \;\;\;\;\bm{\epsilon} \sim \mathcal{N}(\bm{0},\bm{I}),
\end{align*}
Similarly, the probability distributions $q(\bm{x}_t | \bm{x}_0)$ and $\bm{x}_t$ can be described as follows:
\begin{align*}
    q(\bm{x}_t | \bm{x}_0) :&= \mathcal{N}\left(\bm{x}_t; \sqrt{\bar{\alpha}_t} \bm{x}_0, \sqrt{1-\bar{\alpha_t}} \bm{I}\right), \\
    \bm{x}_t &= \sqrt{\bar{\alpha}_t} \bm{x}_0 + \sqrt{1-\bar{\alpha_t}} \bm{\epsilon},
\end{align*}
where $\alpha_t$ and $\bar{\alpha_t}$ are defined as follows:
\begin{align*}
    \alpha_t := 1-\beta_t, \;\;\;\;\bar{\alpha}_t := \prod_{s=1}^t \alpha_s,
\end{align*}

\begin{table*}[t]
\centering
\normalsize
\renewcommand{\arraystretch}{1.2}
\caption{Quantitative results on comparative experiments. ``ME'' refers to the Multimodal Encoder in the intermediate training step.}
\label{tab:result}
\begin{tabular}{l|ccccccccc}
\hline
\multicolumn{1}{c}{\small{[\%]}}                                      & Diff. step                & ME                        & mIoU                                                               & oIoU                                                               & P@0.5                                                              & P@0.6                                                              & P@0.7                                                              & P@0.8                                                             & P@0.9                                                             \\ \hline
(i) LAVT\cite{yang2022lavt} &                           &                           & 24.27\tiny{$\pm{3.15}$}                           & 22.25\tiny{$\pm{2.85}$}                           & 21.27\tiny{$\pm{5.66}$}                           & 13.37\tiny{$\pm{3.74}$}                           & 5.97\tiny{$\pm{2.50}$}                            & 0.94\tiny{$\pm{0.38}$}                           & 0.00\tiny{$\pm{0.00}$}                           \\ \hline
\multirow{2}{*}{(ii) Ours}                   &                           & \checkmark & 30.19\tiny{$\pm{3.98}$}                           & 27.08\tiny{$\pm{2.89}$}                           & 31.66\tiny{$\pm{6.52}$}                           & 23.04\tiny{$\pm{4.66}$}                           & 10.33\tiny{$\pm{1.63}$}                           & 1.55\tiny{$\pm{1.36}$}                           & 0.00\tiny{$\pm{0.00}$}                           \\ \cline{2-10} 
                                            & \checkmark & \checkmark & \textbf{34.40\tiny{$\pm{3.79}$}} & \textbf{31.59\tiny{$\pm{3.03}$}} & \textbf{36.63\tiny{$\pm{6.14}$}} & \textbf{27.79\tiny{$\pm{5.28}$}} & \textbf{16.30\tiny{$\pm{2.98}$}} & \textbf{6.41\tiny{$\pm{1.19}$}} & \textbf{0.66\tiny{$\pm{0.62}$}} \\ \hline
\end{tabular}
\end{table*}

\begin{table*}[]
\centering
\caption{Ablation study under condition(b). ``Feat.'' refers to the features obtained in the intermediate training step.}
\normalsize
\label{tab:ab-2}
\renewcommand{\arraystretch}{1.6}
\begin{adjustwidth}{-1in}{-1in} \begin{center} \resizebox{\textwidth}{!}{
\begin{tabular}{cccccccccc}
\hline
\small{[\%]} & \multicolumn{1}{l}{Cond.} & Feat.                                                                                                
                            & mIoU                & oIoU                & P@0.5               & P@0.6               & P@0.7      & P@0.8              & P@0.9              \\ \hline
     & (b-1)                  & $\bm{H}_{\mathrm{it}}^{(0)}$, $\bm{H}_{\mathrm{it}}^{(1)}$, $\bm{H}_{\mathrm{it}}^{(2)}$, $\bm{H}_{\mathrm{it}}^{(3)}$                                       & 34.09\tiny{$\pm{4.14}$}          & 31.57\tiny{$\pm{3.07}$}          & 35.52\tiny{$\pm{6.04}$}          & 27.29\tiny{$\pm{4.93}$} & \textbf{16.35\tiny{$\pm{3.16}$}} & 6.35\tiny{$\pm{1.22}$}          & \textbf{1.82\tiny{$\pm{1.33}$}} \\
     & (b-2)                  & $\bm{H}_{\mathrm{it}}^{(0)}$, $\bm{H}_{\mathrm{it}}^{(1)}$, $\bm{H}_{\mathrm{it}}^{(2)}$                                       & \textbf{34.40\tiny{$\pm{3.79}$}} & \textbf{31.59\tiny{$\pm{3.03}$}} & \textbf{36.63\tiny{$\pm{6.14}$}} & \textbf{27.79\tiny{$\pm{5.28}$}} & 16.30\tiny{$\pm{2.98}$} & \textbf{6.41\tiny{$\pm{1.19}$}} & 0.66\tiny{$\pm{0.62}$}          \\
     & (b-3)                  & $\bm{H}_{\mathrm{it}}^{(0)}$, $\bm{H}_{\mathrm{it}}^{(1)}$, $\bm{H}_{\mathrm{it}}^{(3)}$                                       & 33.68\tiny{$\pm{4.37}$}          & 30.61\tiny{$\pm{4.17}$}          & 35.80\tiny{$\pm{6.73}$}          & 26.46\tiny{$\pm{4.66}$}          & 15.69\tiny{$\pm{3.88}$} & 6.08\tiny{$\pm{1.58}$}          & 0.50\tiny{$\pm{0.41}$}          \\
     & (b-4)                  & $\bm{H}_{\mathrm{it}}^{(0)}$, $\bm{H}_{\mathrm{it}}^{(2)}$, $\bm{H}_{\mathrm{it}}^{(3)}$                                       & 33.44\tiny{$\pm{4.52}$}          & 30.51\tiny{$\pm{3.89}$}          & 35.03\tiny{$\pm{6.36}$}          & 26.85\tiny{$\pm{6.02}$}          & 15.91\tiny{$\pm{4.36}$} & 5.25\tiny{$\pm{0.82}$}          & 1.66\tiny{$\pm{1.48}$}          \\
     & (b-5)                  & $\bm{H}_{\mathrm{it}}^{(1)}$, $\bm{H}_{\mathrm{it}}^{(2)}$, $\bm{H}_{\mathrm{it}}^{(3)}$                             & 32.54\tiny{$\pm{4.97}$}          & 29.97\tiny{$\pm{4.14}$}          & 35.30\tiny{$\pm{6.72}$}          & 26.24\tiny{$\pm{6.26}$}          & 13.15\tiny{$\pm{3.77}$} & 3.26\tiny{$\pm{1.95}$}          & 0.50\tiny{$\pm{0.41}$}          \\ \hline
\end{tabular}
} \end{center} \end{adjustwidth} 
\end{table*}

\subsubsection{Crossmodal Reverse Decoder}
The Crossmodal Reverse Decoder takes as input ($\bm{x}_t$, $\bm{H}_{\mathrm{it}}^{(i)}$, $p\left(\hat{\bm{y}}_{\mathrm{it}}\right)$) and outputs the predicted probability $p\left(\hat{\bm{y}}_{\mathrm{diff}}\right)$ of being part of the target mask at each pixel of $H \times W$. 

First, we obtain the $t$-th added noise $\bm{\epsilon}_t \in \mathbb{R}^{H \times W \times 3}$ in the Forward Encoder, using $\bm{x}_t$.
$n_b$-th predictive noise $\hat{\bm{\epsilon}}_t^{(n_b)} \in R^{H_{n_b} \times W_{n_b} \times C_{n_b}}$ is extracted by the pre-trained DDPM\cite{ho2020denoising}, where $n_b$ denotes the index number of each layer in the UNet structure, and $H_{n_b}$, $W_{n_b}$, and $C_{n_b}$ denote the height of the image, width of the image, and the number of channels in the $n_b$-th layer, respectively.
We then obtain the multimodal feature $\bm{H}_{\mathrm{seg}}^{\prime(n_b)} \in \mathbb{R}^{H_{n_b} \times W_{n_b} \times C_{n_b}}$ as follows:
\begin{align*}
  \hat{\bm{x}}_{t-1}^{(n_b)} &= \bm{x}_{t}^{(n_b)} - \hat{\bm{\epsilon}}_t \\
  \bm{H}_{\mathrm{seg}}^{\prime(n_b)} &= \hat{\bm{x}}_{0}^{(n_b)} + \bm{H}_{\mathrm{it}}^{(n_b)}
\end{align*}
where $\hat{\bm{x}}_{t}^{(n_b)}$ denotes the predicted value of $\bm{x}_t^{(n_b)}$.
Next, we extract the resized features $\bm{H}_{\mathrm{seg}}^{(n_b)} \in \mathbb{R}^{H_1 \times W_1 \times C_{n_b}}$ using bilinear interpolation on $\bm{H}_{\mathrm{seg}}^{\prime(n_b)}$. 
We obtain the multimodal feature $\bm{H}_{\mathrm{seg}}\in \mathbb{R}^{H \times W \times C_{\mathrm{seg}}}$ by concatenating all the obtained $\bm{H}_{\mathrm{seg}}^{(n_b)}$ in the channel direction, where $C_{\mathrm{seg}}$ denotes the dimension of the feature. 
Finally, we obtain the estimated probability $p\left(\hat{\bm{y}}_{\mathrm{diff}}\right)$ in the diffusion step using $\bm{H}_{\mathrm{seg}}$ and $p\left(\hat{\bm{y}}_{\mathrm{it}}\right)$ as follows:
\begin{align*}
  \Delta p &= f_{\mathrm{BN}}(\mathrm{ReLU}(f_{\mathrm{FC}}(\bm{H}_{\mathrm{seg}}))) \\
  p(\hat{\bm{y}}_{\mathrm{diff}}) &= p\left(\hat{\bm{y}}_{\mathrm{it}}\right) + \Delta p, 
\end{align*}
where $\bm{z}_{\mathrm{diff}}$ denotes the predicted difference between the binary ground-truth mask $\bm{y}$ and $p\left(\hat{\bm{y}}_{\mathrm{it}}\right)$ in $H \times W$.
Additionally, $f_{\mathrm{BN}}$ and $f_{\mathrm{FC}}$ denote batch normalization and linear combination, respectively. Finally, $p(\hat{\bm{y}}_{\mathrm{diff}})$ is binarised at a threshold of 0.5 to output the $H \times W$ binary prediction mask image $\hat{\bm{y}}_{\mathrm{diff}}$.

\subsection{Loss Function}
We use the cross-entropy loss as the loss function in the intermediate training step and the mean absolute error in the diffusion step.

\section{Experiments
\label{exp}
}
\vspace{-1mm}
\subsection{Dataset}
\vspace{-1mm}

In this study, we focuse on generating a segmentation mask for objects in real-world indoor environments based on a natural language instruction.
The REVERIE dataset\cite{qi2020reverie}, which is a standard dataset for VLN and object localization, is closely related to the present study because its images were collected in real-world indoor environments.
However, this dataset does not include pixel-wise mask images of objects. Therefore, we constructed the Segmentation from Household-task Instructions on Manipulation in Real Indoor Environment (SHIMRIE) dataset, which is suitable for the OSMI task by extracting mask images using the Matterport3D\cite{chang2018matterport3d}, which contains voxel-wise class information. 

The OSMI task requires instructions and images of household tasks that manipulate the target object, in addition to masking images of the target object. However, to best of our knowledge, no existing dataset satisfies these requirements.
Therefore, we constructed the SHIMRIE dataset, which satisfies all the requirements.

We collected mask images using voxel-wise class information contained in the Matterport3D and the bounding box of the target object contained in the REVERIE dataset as follows:
First, we assigned a color to each class and painted each voxel for all floor maps. 
Then, we obtained the camera parameters for each image in the REVERIE dataset and performed the same transformation.
Next, we extracted the 2D images painted for each class.
Finally, we collected the ground-truth mask image that had its largest segment contained within its bounding box region as the mask of the target object. 
In parallel, we collected instructions from the REVERIE dataset that corresponded to the ground-truth mask images.
In this study, we resized images with a resolution of 640$\times$480 to 256$\times$256 to reduce computational cost.
In the study of \cite{qi2020reverie}, the annotators were asked to provide instructions according to randomly selected objects and paths.
In the study of \cite{chang2018matterport3d}, labels were provided for each object using a crowdsourcing service.

\begin{figure*}[t]
        \centering
        \begin{tabular}{c}
            \begin{minipage}{0.3\hsize}
                \centering
                \includegraphics[width=\linewidth]{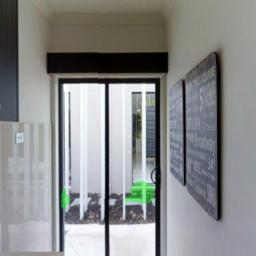}
            \end{minipage}
            \begin{minipage}{0.3\hsize}
                \centering
                \includegraphics[width=\linewidth]{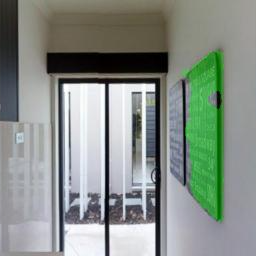}
            \end{minipage}
            \begin{minipage}{0.3\hsize}
                \centering
                \includegraphics[width=\linewidth]{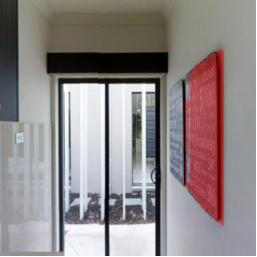}
            \end{minipage} \\
            ``Go to the laundry room and straighten the picture closest  to the light switch.'' \\
            \begin{minipage}{0.3\hsize}
                \vspace{2mm}
                \centering
                \includegraphics[width=\linewidth]{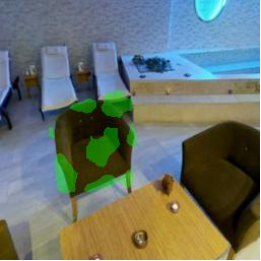}
            \end{minipage}
            \begin{minipage}{0.3\hsize}
                \vspace{2mm}
                \centering
                \includegraphics[width=\linewidth]{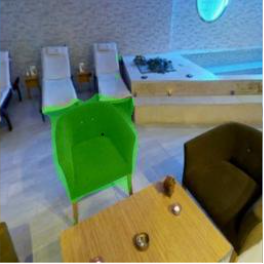}
            \end{minipage}
            \begin{minipage}{0.3\hsize}
                \vspace{2mm}
                \centering
                \includegraphics[width=\linewidth]{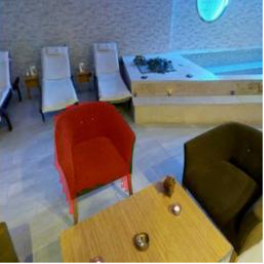}
            \end{minipage} \\
            ``Go to the lounge and remove the small brown chair facing the counter.'' \\
            \begin{minipage}{0.3\hsize}
                \centering
                \small (a) Baseline\cite{yang2022lavt}
            \end{minipage}
            \begin{minipage}{0.3\hsize}
                \centering
                \small (b) Ours
            \end{minipage}
            \begin{minipage}{0.3\hsize}
                \centering
                \small (c) Ground-truth
            \end{minipage}
        \end{tabular}
        \caption{Qualitative result of successful cases. The baseline method incorrectly masked irrelevant outdoor areas; however, the proposed method generated a more appropriate mask for the target object than the baseline method.}
        \label{fig:success}
\end{figure*}

The SHIMRIE dataset contained 4341 images and 11371 sentences, with a vocabulary size of 3558 words, a total of 196541 words, and an average sentence length of 18.8 words.
In the SHIMRIE dataset, the training, validation, and test sets consisted of 10153; 856; and 362 samples, respectively. 
We collected the images for this dataset from 90 floor maps, which we split into seen and unseen environments according to the split defined in the REVERIE dataset.
The training set contained samples only from the seen environments.
The validation set contained 582 and 274 samples from the seen and unseen environments, respectively. 
The test set contained samples only from the unseen environments.

In the intermediate training step, we used the training set to train the model, the validation set to tune the hyperparameters, and the test set to evaluate the performance of the model.
In the diffusion step, we used the validation set to train the model and the test set to evaluate the performance of the model.

\begin{table}[b]
\centering
\renewcommand{\arraystretch}{1.2}
\normalsize
\caption{Parameter settings for the MDSM.}
\label{tab:hyperparameter}
\begin{tabular}{ccc}
\hline
\multicolumn{1}{l}{} & Intermediate. step             & Diffusion step                        \\ \hline
Epoch                & 11                                     & 5                                     \\ \hline
Batch Size           & 16                                     & 1                                     \\ \hline
\begin{tabular}[c]{@{}c@{}}Learning \\ Rate\end{tabular}        & $5 \times 10^{-5}$                            & $1 \times 10^{-3}$                           \\ \hline
Optimizer     & \begin{tabular}[c]{@{}c@{}}AdamW \\ \small{$\left(\beta_1=0.9, \beta_2=0.99\right)$}\end{tabular} & \begin{tabular}[c]{@{}c@{}}Adam\\ \small{$\left(\beta_1=0.9, \beta_2=0.99\right)$}\end{tabular} \\ \hline
\end{tabular}
\end{table}

\vspace{-1mm}
\subsection{Experimental Setup}
\vspace{-0.8mm}
Table~\ref{tab:hyperparameter} shows the hyperparameter settings of the proposed method.
The total number of trainable parameters in the intermediate training step and diffusion step were 123M and 1.09M, respectively. The total number of multiply-add operations in the steps were 508G and 71.2G, respectively.
Our model was trained on a GeForce RTX 3090 with 24GB of memory and an Intel Core i9-10900KF with 64GB of memory.
The training time for the proposed method was approximately 1 hour and 45 minutes for the intermediate training step and 1 hour and 25 minutes for the diffusion step.
The inference time per sample was approximately 0.039 seconds for the intermediate training step and 0.36 seconds for the diffusion step.

We considered the final performance as the evaluation on the test set when the validation set loss was the minimum in the diffusion step.
If there was no improvement for 50 consecutive iterations, we terminated training. 
Here, we considered the first three epochs as the warming-up epochs.

\vspace{-1mm}
\subsection{Quantitative Results}
\vspace{-0.8mm}
Table~\ref{tab:result} shows the quantitative results for comparing the baseline method (i) and the variants of the proposed method. The results show the means and standard deviations for the five experiments. We also experimented with the proposed method in two cases (ii) and (iii): without and with the diffusion step, respectively.
We selected LAVT\cite{yang2022lavt} because it was successfully applied to the RES task, which is closely related to OSMI task.

As evaluation metrics, we used mIoU, oIoU, and precision at the 0.5, 0.6, 0.7, 0.8 and 0.9 threshold values.
These are standard metrics for the RES task, which is closely related to the OSMI task.
The primary metric is mIoU because it treats large and small objects equally, whereas oIoU favors large objects.
The metrics mIoU , oIoU, and Precision@$k$ (P@$k$)) are defined as follows:
\begin{align*}
  \mathrm{mIoU}(\hat{\bm{y}}, \bm{y}) &= \frac{1}{N} \sum_{i=1}^N \mathrm{IoU}(\hat{\bm{y}}_i , \bm{y}_i), \\
  \mathrm{oIoU}(\hat{\bm{y}}, \bm{y}) &= \frac{\sum_{i=0}^N\left(\hat{\bm{y}}_i \cap  \bm{y}_i\right)}{\sum_{i=0}^N\left(\hat{\bm{y}}_i \cup  \bm{y}_i\right)}, \\
  \mathrm{P@}k &=\frac{T_k}{N},
\end{align*}
where $\hat{\bm{y}}$ and $\bm{y}$ denote the predicted and  ground-truth segmentation masks, respectively. $N$ denotes the number of samples. $\hat{\bm{y}}_i$ and $\bm{y}_i$ denote the predicted and ground-truth segmentation masks in the $i$-th sample, respectively. $T_k$ denotes the number of samples for which the IoU between the predicted mask and ground-truth mask exceeds the threshold $k$.

Table \ref{tab:result} shows that the mIoU for method (i) and method (ii) were 24.27 and 30.19 \%, respectively.
Therefore, the result of method (ii) was 5.92 points higher than that of method (i). Additionally, method (ii) performed as well as or better than method (i) for both oIoU and $\mathrm{P@}k$. 
The mIoU for method (iii) was 34.40 \%, which was 10.13 points higher than that for method (i). Additionally, method (iii) outperformed method (i) for both oIoU and $\mathrm{P@}k$. 
Therefore, method (iii) performed best out of the three methods.
The difference in performance between method (i) and method (iii) was statistically significant ($p$-value < 0.05).

\vspace{-1mm}
\subsection{Qualitative Results}
\vspace{-0.8mm}
Fig.~\ref{fig:success} shows the qualitative results.
In the figure, panels (a), (b), and (c) show the prediction result of the baseline, the prediction result of our proposed method, and the ground-truth, respectively.
The given instruction for the upper images was ``Go to the laundry room and straighten the picture closest to the light switch.'' The target object was the painting masked in panel (c).
The given instruction for the lower images was ``Go to the lounge and remove the small brown chair facing the counter.'' The target object was the chair masked in panel (c).

\begin{figure*}[t]
        \centering
        \begin{tabular}{c}
            \begin{minipage}{0.3\hsize}
                \centering
                \includegraphics[width=\linewidth]{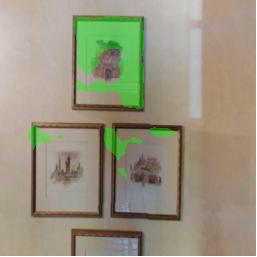}
            \end{minipage}
            \begin{minipage}{0.3\hsize}
                \centering
                \includegraphics[width=\linewidth]{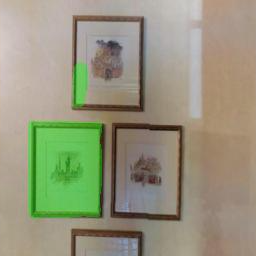}
            \end{minipage}
            \begin{minipage}{0.3\hsize}
                \centering
                \includegraphics[width=\linewidth]{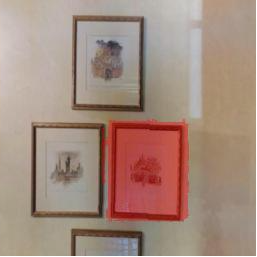}
            \end{minipage} \\
            ``Visit the bathroom and bring me the picture nearest the toilet.'' \\
            \begin{minipage}{0.3\hsize}
                \centering
                \small (a) Baseline
            \end{minipage}
            \begin{minipage}{0.3\hsize}
                \centering
                \small (b) Ours
            \end{minipage}
            \begin{minipage}{0.3\hsize}
                \centering
                \small (c) Ground-truth
            \end{minipage}
        \end{tabular}
        \caption{Qualitative result of the failure case. (a) LAVT\cite{yang2022lavt},  (b) our method (MDSM), and (c) the ground-truth mask.}
        \label{fig:failure}
\end{figure*}

Fig.~\ref{fig:failure} shows the qualitative result of a failure case.
The given instruction sentence was ``Visit the bathroom and bring me the picture nearest the toilet.'' The target object was the painting masked in panel (c).
The baseline and our method incorrectly masked paintings that were different from the target object.
This is presumably because the toilet was outside the input image.

\vspace{-1mm}
\subsection{Ablation Studies}
\vspace{-0.8mm}

We set the following ablation conditions:
\begin{enumerate}
    \renewcommand{\labelenumi}{(\alph{enumi})}
    \item W/o the crossmodal parallel feature extraction: We removed crossmodal parallel feature extraction to investigate its effect on performance in the intermediate training step.
    \item Selective $\bm{H}_{\mathrm{it}}^{(i)}$ in the Crossmodal Reverse Decoder: In the diffusion step, we investigated the effect of each $\bm{H}_{\mathrm{it}}^{(i)}$ on performance.
\end{enumerate}

Tables \ref{tab:result} and \ref{tab:ab-2} show the quantitative results of the ablation studies under conditions (a) and (b), respectively.
Table \ref{tab:result} shows that the mIoU was 5.92 points lower without the crossmodal parallel feature extraction mechanism.
Additionally, oIoU and $\mathrm{P@}k$ were significantly worse, except for P@0.9. These results indicate that the crossmodal parallel feature extraction mechanism contributed to performance.
Table \ref{tab:ab-2} shows that the mIoUs under conditions (b-3), (b-4) and (b-5) were lower than that under condition (b-1) by 1.55, 0.65 and 0.41 points, respectively.
In contrast, the mIoU in condition (b-2) was 0.31 points higher than in condition (b-1).
This indicates that $\bm{H}_{\mathrm{it}}^{(0)}$ contributed the most to the performance improvement.
We believe that this is because the fine-grained features contained in $\bm{H}_{\mathrm{it}}^{(0)}$ were effective for pixel-wise segmentation.

\vspace{-1mm}
\subsection{Error Analysis}
\vspace{-0.8mm}

In the test set, there were 25 samples for which $\mathrm{IoU}(\hat{\bm{y}}, \bm{y}) = 0$.
\begin{table}[b]
\centering
\caption{Error analysis.}
\label{tab:errors}
\renewcommand{\arraystretch}{1.3}
\begin{tabular}{clc}
\hline
Errors & \multicolumn{1}{c}{Description}                                                                                         & \#Error \\ \hline
SC     & \begin{tabular}[c]{@{}l@{}}Serious comprehension errors for handling \\ visual and language information\end{tabular} & 11      \\ \hline
RE     & \begin{tabular}[c]{@{}l@{}}Reference/Exophora resolution errors for \\ linguistic information\end{tabular}              & 31      \\ \hline
BTE    & Segmentation of extra objects                                                                                           & 19      \\ \hline
WNS    & Over- or under-segmentation                                                                                             & 16      \\ \hline
NSG    & \begin{tabular}[c]{@{}l@{}}No segmentation in any region of the image\end{tabular}                                    & 11      \\ \hline
SNI    & \begin{tabular}[c]{@{}l@{}}Segmentation of non-target objects in the \\ instruction\end{tabular}                         & 6       \\ \hline
AE     & \begin{tabular}[c]{@{}l@{}}Annotation errors in the ground-truth mask \\ or instruction\end{tabular}                    & 6       \\ \hline
Total  & -                                                                                                                       & 100     \\ \hline
\end{tabular}
\end{table}
Table \ref{tab:errors} describes the categories of failure cases. We analyzed the worst 100 samples that have low IoU values.
We can roughly divide the causes of failure into seven types:

\subsubsection{Serious comprehension error (SC)}
The SC category refers to failure cases in which our model failed to handle visual and linguistic information; 
that is, the model incorrectly generated a mask for an object that was not contained in the instruction.
For example, an incorrect segmentation mask was generated for a ``desk’’ given the instruction ``Fetch me a pillow.''

\subsubsection{Reference/exophora resolution error (RE)}
The RE category refers to failure cases in which our model failed to understand the referring expressions correctly.
For example, an incorrect segmentation mask for ``the pillow on the left-hand side’’ was generated given the instruction ``Fetch the pillow farthest to the right of the three pillows.''

\subsubsection{Segmentation of extra objects (SEO)}
The SEO category refers to cases in which our model masked non-target objects in addition to the target object.
For example, incorrect segmentation masks were generated for ``both pillows on the left-hand and right-hand side’’ given the instruction ``Fetch the pillow farthest to the right of the three pillows.''

\subsubsection{Over- or under-segmentation (OUS)}
The OUS category refers to over or under-segmentation.

\subsubsection{No segmentation generated error (NSG)}
The NSG category refers to cases in which our model masked no regions in the image.

\subsubsection{Segmentation of non-target objects in the instruction (SNI)}
The SNI category refers to cases in which our model masked non-target objects in the instruction.
For example, an incorrect segmentation mask was generated for a ``bed’’ given the instruction ``Fetch me a pillow on the bed.''

\subsubsection{Annotation error (AE)}
The AE category refers to annotation errors in the ground-truth mask or instruction.
For example, the ground-truth mask contained non-target objects, or the instruction was grammatically incorrect.

Table \ref{tab:errors} indicates that the main bottleneck was RE.
A possible solution to this problem is to use the left and right contextual images to understand the locations of objects in 3D space or use syntactic analyzers to decompose input sentences into multiple referring expressions.

\vspace{-1mm}
\section{Conclusions}
\vspace{-1mm}

In this study, we focused on the OSMI task and generated a segmentation mask for a target object of instructions for household tasks.
We emphasize the following contributions of this study:

\begin{itemize}
    \item We proposed MDSM, which is a two-stage multimodal segmentation model.
    \item We introduced a crossmodal parallel feature extraction mechanism using CLIP\cite{radford2021learning} and Swin Transformer\cite{liu2021swin}.
    \item We extended DDPM\cite{ho2020denoising} to handle crossmodal features in the novel Crossmodal Reverse Decoder.
    \item The MDSM outperformed the baseline method on the SHIMRIE dataset, which we newly constructed for the OSMI task.
\end{itemize}

In future work, we plan to introduce this model to physical robots.


\section*{ACKNOWLEDGMENT}

This work was partially supported by JSPS KAKENHI Grant Number 20H04269, JST Moonshot, and NEDO.
\bibliographystyle{IEEEtran}
\bibliography{reference}

\end{document}